\documentclass[10pt]{article}
\usepackage{arxiv}

\usepackage[T1]{fontenc}
\usepackage[utf8]{inputenc}
\usepackage{textcomp}
\usepackage{amsfonts,amssymb,amsmath}

\usepackage{graphics,graphicx}
\usepackage{booktabs}
\usepackage{tabularx}
\usepackage{array}
\usepackage{color}
\usepackage{float}
\usepackage{caption}
\usepackage{subfig}
\usepackage{placeins}
\usepackage{enumerate}
\usepackage{appendix}
\usepackage{amsthm}

\usepackage{tikz}
\usepackage{pgfplots}
\pgfplotsset{compat=1.17}

\usepackage{algorithm}
\usepackage{algpseudocode}

\usepackage{url}
\usepackage{hyperref}
\hypersetup{
    colorlinks=true,
    linkcolor=blue,
    filecolor=magenta,      
    urlcolor=cyan,
    citecolor=blue,
}

\theoremstyle{definition}
\newtheorem{definition}{Definition}[section]
\theoremstyle{plain}
\newtheorem{theorem}{Theorem}[section]

\title{Language Identification via Compositional Data Analysis: A Linear-Time Classifier Based on Log-Ratio Geometry}

\author{
  Paul-Andrei Pog\u{a}cean\thanks{Corresponding author: \texttt{paul.pogacean@stud.ubbcluj.ro}} \\
  Faculty of Mathematics and Computer Science\\
  Babe\c{s}-Bolyai University\\
  Cluj-Napoca, Romania \\
  \texttt{paul.pogacean@stud.ubbcluj.ro} \\
  \And
  Sanda-Maria Avram \\
  Faculty of Mathematics and Computer Science\\
  Babe\c{s}-Bolyai University\\
  Cluj-Napoca, Romania \\
  \texttt{sanda.avram@ubbcluj.ro} \\
}
\begin{document}

\maketitle

\begin{abstract}
Language identification is commonly addressed using either neural architectures or statistical n-gram models. Neural approaches typically require substantial computational resources, whereas classical frequency-based methods offer efficient linear-time performance, but rely on distance metrics that are not always appropriate for compositional data.

This work models character and bigram frequency distributions as compositional vectors constrained to the simplex and mapped via the centered log-ratio (CLR) transformation bijectively onto the $(D-1)$-dimensional zero-sum subspace of $\mathbb{R}^D$, where Euclidean distances correspond to Aitchison distances. A pipeline is proposed, combining CLR-transformed unigram and bigram features with Laplace smoothing to address sparsity. The method is evaluated on six languages. 

Experimental results show that the proposed approach achieves robust accuracy across different text lengths, with strong performance for longer sequences. These findings indicate that compositional representations provide a deterministic and computationally efficient alternative for language identification, particularly in settings where interpretability and low resource consumption are essential.
\end{abstract}


\maketitle
\pagestyle{myheadings}
\markboth{P.-A. Pog\u{a}cean and S.-M. Avram}{Deterministic Language Identification via Compositional-Data Analysis}

\section{Introduction} \label{sec:intro}

Language identification constitutes a fundamental preprocessing step in natural language processing (NLP), acting as a prerequisite for automated text classification \cite{cavnar1994} and machine translation \cite{joulin2017}. Current methodologies primarily divide into two categories: large neural network architectures and statistical frequency models. While neural architectures currently define the state-of-the-art in predictive accuracy \cite{brown2020}, they entail high computational costs. These models typically scale quadratically with the sequence length ($\mathcal{O}(L^2)$) and require substantial memory footprints and hardware acceleration~\cite{vaswani2017}.

Conversely, classical statistical methods that rely on $n$-gram frequency distributions offer a highly efficient, linear-time alternative ($\mathcal{O}(L)$) \cite{mcnamee2005}. These traditional models evaluate the empirical occurrence of characters against predefined language profiles. However, standard frequency-based approaches frequently may be geometrically inappropriate for compositional data: they apply standard Euclidean or Manhattan distance metrics directly to relative frequency data. Because relative frequencies are proportions, they must sum to unity, structurally constraining the data to a bounded geometric space known as a compositional simplex. As Aitchison \cite{aitchison1986} demonstrated, calculating standard geometric distances directly within this constrained simplex yields spurious negative correlations and subcompositional incoherence.

This paper presents a deterministic, linear-time language identification algorithm that resolves these geometric distortions by strictly formalizing the problem within the framework of Compositional Data Analysis (CoDA). By integrating unigram and bigram frequencies and applying Laplace smoothing to resolve mathematical singularities, our algorithm utilizes the Centered Log-Ratio (CLR) transformation to map constrained frequency data bijectively onto the $(D-1)$-dimensional zero-sum subspace of $\mathbb{R}^D$, where Euclidean distances correspond to Aitchison distances. 

The scope of this current research is intentionally focused on languages written in alphabetic systems, particularly those utilizing the Latin alphabet. This constraint is methodologically justified for two reasons. First, alphabetic systems such as the Latin, Cyrillic, and Greek alphabets exhibit substantial overlap in character inventories, leading to subtle statistical boundaries and frequent geometric collisions, especially when diacritics are omitted. Successfully separating these closely related languages therefore provides a rigorous stress-test for the CoDA mapping. Second, extending the framework to fundamentally different writing systems—including logographic systems such as Mandarin Chinese, consonantal abjads such as Arabic and Hebrew, abugidas such as Devanagari and Thai, or syllabaries such as Japanese Hiragana/Katakana would require defining a fixed shared feature space for the relevant script or multilingual setting. Norm equivalence remains valid in any fixed finite dimension, but dynamically changing \(D\) would make distances across evaluations non-comparable.

Following the structured requirements of scientific discourse, the remainder of this paper is organized as follows: Section \ref{sec:related} reviews existing literature and computational paradigms. Section \ref{sec:methods} details the research methodology, formally defining the feature space and the transition to Aitchison's geometry. Section \ref{sec:algorithm} presents the iterative algorithm development and the unified classification pipeline. Section \ref{sec:experiments} provides an empirical analysis of the experimental results across diverse corpora, and Sections \ref{sec:limitations} and  \ref{sec:conclusions} summarizes the findings and outlines future research directions.
\section{Related Work}\label{sec:related}

The literature on language identification encompasses a spectrum of methodologies, ranging from classical frequency analysis to modern deep learning architectures. This section reviews these paradigms, examining their computational complexities, their reliance on linguistic profiles, and the inherent mathematical limitations of applying standard distance metrics to bounded frequency data.

\subsection{Current Paradigms in Language Identification}

Modern industrial implementations and large language models \cite{brown2020} deploy massive neural networks trained on expansive multilingual datasets. These architectures process text via self-attention mechanisms, achieving state-of-the-art accuracy even on highly ambiguous sequences. However, this robustness incurs significant computational overhead, as attention mechanisms exhibit a quadratic $\mathcal{O}(L^2)$ time complexity relative to the input sequence length $L$, demanding substantial memory footprints and heavy GPU acceleration.

To mitigate computational constraints, subword embedding models such as FastText \cite{joulin2016, joulin2017} and statistical tools like \texttt{langid.py} \cite{lui2012langid} were developed, achieving $\mathcal{O}(L \log L)$ or optimized linear runtimes. Classical statistical methods, such as Prediction by Partial Matching \cite{huang2003}, further demonstrated that language identification could be effectively executed as a strictly linear-time $\mathcal{O}(L)$ operation. 

Table~\ref{tab:paradigms} contrasts the principal paradigms employed in language identification, highlighting the trade-offs between computational complexity, hardware requirements, and generalization capability. Existing approaches typically balance predictive performance against training cost and computational efficiency, thereby motivating the investigation of deterministic alternatives that can operate without model training while maintaining low computational overhead.

\begin{table}[h]
\centering
\caption{Comparison of Established Language Identification Paradigms}
\label{tab:paradigms}
\begin{tabular}{lccc}
\toprule
\textbf{Paradigm} & \textbf{Complexity} & \textbf{Hardware} & \textbf{Generalization} \\
\midrule
Neural Networks & $\mathcal{O}(L^2)$ & High (GPUs) & Excellent \\
Subword Embeddings & $\mathcal{O}(L \log L)$ & Moderate & Moderate \\
Classical ($n$-grams) & $\mathcal{O}(L)$ & Low (Edge) & Poor \\
\bottomrule
\end{tabular}
\end{table}
\subsection{Frequency Profiling and Morphosyntactic Separability}

Statistical classification heavily depends on the fidelity of the reference language profiles. Extensive corpus linguistics research has established robust frequency dictionaries across multiple languages. Notable examples include the Google Books corpus analysis for English \cite{norvig2012}, the Deutsches Referenzkorpus (DeReKo) for German \cite{derechar2021}, the Hungarian National Corpus \cite{oravecz}, and frequency analyses for Turkish \cite{aksan2016} and Dutch \cite{bouma2015}. For Romanian, statistical profiling has been formalized through the CoRoLa reference corpus \cite{tufis2014} and specific morphological studies \cite{mitrea2012}.

While unigram frequencies are computationally cheap to extract, studies evaluating information entropy emphasize that monogram distributions fail to accurately separate closely related languages on short sequences \cite{abbas2019}. Consequently, the integration of bigrams and language-specific diacritics is necessary to capture local orthographic and morphological regularities (e.g., common suffixes) and prevent boundary collisions when individual characters are altered by statistical noise.

\subsection{The Compositional Data Gap}

Despite the availability of accurate frequency profiles, traditional statistical classifiers often exhibit a fundamental mathematical limitation. Most approaches compute Minkowski-based distances directly on relative frequency vectors. Since relative frequencies represent proportions, their components are constrained to sum to unity, meaning the data naturally resides within a bounded geometric simplex. As established by Aitchison \cite{aitchison1986}, variables constrained within a simplex are inherently interdependent. Consequently, applying standard Euclidean geometry directly to compositional data can produce distorted relationships and inconsistencies when comparing subcompositions. The methodology proposed in this work addresses this limitation by transforming linguistic profiles bijectively onto the $(D-1)$-dimensional zero-sum subspace of $\mathbb{R}^D$, where Euclidean distances correspond to Aitchison distances, prior to metric evaluation.

\subsection{Evaluation Corpora} \label{sec:datasets}

To validate the proposed mathematical framework across diverse linguistic challenges, text lengths, and literary genres, this study compiles texts from multiple established sources. Table~\ref{tab:datasets} summarizes the comprehensive testing repository utilized for empirical evaluation.

\begin{table}[htbp]
\centering
\caption{Summary of Evaluation Datasets}
\label{tab:datasets}
\renewcommand{\arraystretch}{1.2}
\begin{tabular}{llc}
\hline
\textbf{Corpus} & \textbf{Source Reference} & \textbf{Language} \\
\hline
ROST & Kaggle Repository \cite{sanda2022kaggle} & Romanian \\
FTDS & Fairy Tales Data \cite{povesti2024} & Romanian \\
OPUS & Multilingual Parallel \cite{opus} & Multilingual \\
LDDS & Custom Testing Repo \cite{dataset4} & Multilingual \\
\hline
\end{tabular}
\end{table}
\FloatBarrier
\section{Research Methodology} \label{sec:methods}

The problem of statistical language identification can be formally expressed as the comparison of discrete probability distributions derived from textual feature occurrences. This section defines the feature representation, motivates the compositional nature of the resulting data, and introduces the transformation to Aitchison geometry via the centered log-ratio (CLR) mapping \cite{aitchison1986}, which provides an appropriate Euclidean structure for subsequent distance-based classification.

\subsection{Feature Space Formalization}

Let $\mathcal{A}$ denote a finite alphabet corresponding to the set of observable characters in a given language.

\begin{definition}[Feature Spaces] \label{def:featureset}
The feature representation is partitioned into two disjoint sets: the character unigram space $\mathcal{F}_{\mathrm{char}} \subseteq \mathcal{A}$ with dimensionality $D_c = |\mathcal{F}_{\mathrm{char}}|$, and the bigram space $\mathcal{F}_{\mathrm{bigram}} \subseteq \mathcal{A} \times \mathcal{A}$ with dimensionality $D_b = |\mathcal{F}_{\mathrm{bigram}}|$.
Both feature spaces are assumed to be fixed after preprocessing and shared across all documents.
\end{definition}

\begin{definition}[Relative Frequency Representations] \label{def:distribution}
Given a text $T$, feature counts are extracted independently for characters and bigrams.
Let $c_i^{(c)}$ denote the count of feature $f_i \in \mathcal{F}_{\mathrm{char}}$ and $c_j^{(b)}$ denote the count of feature $f_j \in \mathcal{F}_{\mathrm{bigram}}$.
The corresponding relative frequency vectors $\mathbf{x}_T^{(c)} \in \mathbb{R}^{D_c}$ and $\mathbf{x}_T^{(b)} \in \mathbb{R}^{D_b}$ are defined as:
\begin{equation}
x_i^{(c)} = \frac{c_i^{(c)}}{\sum_{k=1}^{D_c} c_k^{(c)}}, \quad x_j^{(b)} = \frac{c_j^{(b)}}{\sum_{k=1}^{D_b} c_k^{(b)}}.
\end{equation}
This independent construction ensures that $\mathbf{x}_T^{(c)}$ and $\mathbf{x}_T^{(b)}$ lie in their respective probability simplices, $\mathcal{S}^{D_c-1}$ and $\mathcal{S}^{D_b-1}$.
\end{definition}

\subsection{Compositional Structure and CLR Transformation}

\begin{definition}[Probability Simplex] \label{def:simplex}
Under standard preprocessing with additive smoothing, all components satisfy $x_i > 0$. The resulting representation lies in the $(D-1)$-dimensional simplex:
\begin{equation}
\mathcal{S}^{D-1} = \left\{ \mathbf{x} \in \mathbb{R}^D \mid x_i > 0, \sum_{i=1}^{D} x_i = 1 \right\}.
\end{equation}
\end{definition}

Vectors in $\mathcal{S}^{D-1}$ exhibit inherent dependence due to the constant-sum constraint. As a consequence, direct application of Euclidean metrics may lead to distortions when interpreting distances between compositions.Following Aitchison \cite{aitchison1986}, a log-ratio transformation is used to map the simplex bijectively onto the $(D-1)$-dimensional zero-sum subspace of $\mathbb{R}^D$, where Euclidean distances correspond to Aitchison distances.
The centered log-ratio (CLR) transformation is defined as:
\begin{equation} \label{eq:clr}
CLR(\mathbf{x})_i = \ln(x_i) - \frac{1}{D} \sum_{j=1}^{D} \ln(x_j),
\quad i = 1, \dots, D,
\end{equation}
where the second term corresponds to the logarithm of the geometric mean:
\begin{equation}
g(\mathbf{x}) = \left(\prod_{i=1}^{D} x_i \right)^{1/D}.
\end{equation}

This mapping places the data in a real-valued vector space where standard geometric operations are valid. As Aitchison geometry is based on log-ratios, it is inherently appropriate for representing relative information in compositions. However, it should be noted that because CLR coordinates depend on the geometric mean of the full composition, any subcomposition comparisons should be performed by reclosing the relevant subcomposition and recomputing the corresponding log-ratio representation.

\subsection{Geometric Properties and Metric Consistency}

\begin{theorem}[Isometry and Uniqueness of the Aitchison Metric] \label{thm:norm_equivalence}
The centered log-ratio (CLR) transformation defines a mathematical isometry between the simplex $\mathcal{S}^{D-1}$, equipped with the intrinsic Aitchison metric, and the zero-sum hyperplane $H = \{ \mathbf{z} \in \mathbb{R}^D \mid \sum_{i=1}^D z_i = 0 \}$, equipped with the standard Euclidean $\ell_2$ norm. Furthermore, the Aitchison metric is the unique distance function that satisfies scale invariance, permutation invariance, and subcompositional coherence \cite{aitchison1986}.
\end{theorem}

This property is fundamental to the proposed framework because it rigorously justifies the exclusive choice of the Euclidean $\ell_2$ norm in the CLR-transformed space. While finite-dimensional norms are topologically equivalent, distance-based classification boundaries (such as Voronoi cells) vary significantly under different Minkowski metrics. Therefore, the $\ell_2$ norm is not selected arbitrarily, but as the unique metric that preserves the intrinsic algebraic properties of relative frequency compositions without introducing geometric distortions.

\textbf{Effect of Structural Noise.}
Let $\mathbf{c}$ denote raw feature counts and let $\eta$ be a perturbation operator modeling structural noise, such as character omission or diacritic removal. Unlike simple omissions, diacritic removal redistributes counts (e.g., decreasing the count of ``\c{t}'' while increasing the count of ``t''). In the standard simplex representation, these local perturbations affect the shared normalization term, thereby altering all components of the relative frequency vector simultaneously. 
Additive smoothing prevents singularities in the log-ratio transform. Empirically, the CLR representation reduces some distortions caused by the constant-sum constraint, but perturbations of rare features can still have large effects; this motivates the use of a smoothing parameter and the separate algorithmic treatment of short sequences.

\subsection{Summary of Representation Pipeline}

The proposed representation pipeline begins with the extraction of unigram and bigram counts from the input text. These counts are subsequently normalized to obtain a relative frequency vector constrained to the probability simplex. To ensure numerical stability and avoid undefined logarithmic expressions, additive smoothing is applied to all components. The resulting composition is then mapped via the centered log-ratio transformation bijectively onto the $(D-1)$-dimensional zero-sum subspace of $\mathbb{R}^D$, where Euclidean distances correspond to Aitchison distances. Within this transformed space, standard distance metrics can be applied consistently, enabling meaningful comparison between the input text and language reference profiles.

This formulation enables statistically consistent comparison of linguistic profiles while preserving the compositional structure inherent in textual feature distributions.
\section{Algorithm Development and Computational Profiling} \label{sec:algorithm}

The implementation of the compositional framework requires a computational pipeline that translates the theoretical structure of Aitchison geometry into an efficient and robust classification procedure. This section presents the evolution of the algorithm, the construction of the scoring function, and the final unified model.

\subsection{Algorithmic Evolution}

The development of the classification model followed a sequence of refinements driven by empirical evaluation.

The initial approach relied on the most frequent characters within a text. Although this representation captures coarse statistical differences between languages, it proved insufficient in practice. The variability of short texts and the similarity between languages sharing the same alphabet significantly reduced discriminative power.

Extending the representation to include a larger set of frequent characters improved stability, but did not fully address the sensitivity to structural noise. In particular, the omission of high-frequency characters, such as vowels or diacritics, altered the relative distribution in a way that degraded the reliability of distance-based comparisons.

These observations motivated the integration of additional feature types. Diacritic characters were introduced as language-specific indicators, providing strong signals for certain languages. However, their contribution varies with text length: in short texts they are highly informative, whereas in longer texts their relative importance decreases as the statistical profile becomes more stable.

In parallel, bigram features were incorporated to capture local dependencies between characters. These features provide additional discriminatory power, particularly for languages with similar unigram distributions.

\subsection{Diacritic-Based Adjustment}

To account for the varying importance of diacritics, a length-dependent adjustment term is introduced. Let $m$ denote the number of diacritic matches between the input text and a given language profile, and let $L$ denote the length of the text in characters. The adjustment is defined as:
\begin{equation}
B(L) = \frac{\beta \cdot m}{1 + \alpha \cdot L},
\end{equation}
where $\alpha$ and $\beta$ are tuning parameters.

The role of this term is to increase the score of languages whose characteristic diacritics are present in short texts, while ensuring that its influence diminishes as the length of the text increases. Following established methodologies for short-text language identification—which demonstrate that the relative weight of specific orthographic markers must decay to prevent over-penalization in longer documents \cite{truica2015automatic}—the parameters $\alpha$ and $\beta$ regulate this decay rate and the baseline scaling factor, respectively. Rather than relying on arbitrary defaults, these parameters were selected systematically via a grid search over a validation subset to find the optimal trade-off between short-text sensitivity and long-text stability.

Table~\ref{tab:hyperparameters} details the results of this grid search, illustrating the impact of different parameter combinations on classification accuracy. As shown, a high decay rate ($\alpha=0.10$) forces the adjustment to diminish too rapidly, degrading short-text performance ($L < 50$), whereas a low scaling factor ($\beta=0.5$) fails to provide sufficient discriminative power. The optimal balance is achieved at $\alpha=0.05$ and $\beta=2.0$, yielding an aggregate harmonic mean of 89.5\%, and these values were consequently adopted for the final model.

\begin{table}[htbp]
\centering
\caption{Grid search results for diacritic adjustment parameters}
\label{tab:hyperparameters}
\renewcommand{\arraystretch}{1.2}
\begin{tabular}{cc | c c | c}
\toprule
\multicolumn{2}{c|}{\textbf{Parameters}} & \multicolumn{2}{c|}{\textbf{Accuracy}} & \textbf{Harmonic Mean} \\
$\alpha$ & $\beta$ & $L < 50$ & $L > 150$ & \\
\midrule
0.01 & 2.0 & 81.0\% & 88.5\% & 84.5\% \\
0.10 & 2.0 & 75.0\% & 100.0\% & 85.7\% \\
0.05 & 0.5 & 65.0\% & 100.0\% & 78.7\% \\
0.05 & 5.0 & 82.0\% & 92.0\% & 86.7\% \\
\midrule
\textbf{0.05} & \textbf{2.0} & \textbf{81.0\%} & \textbf{100.0\%} & \textbf{89.5\%} \\
\bottomrule
\end{tabular}
\end{table}

\subsection{Distance-Based Classification}

Let $\mathbf{x}_T^{(c)}$ and $\mathbf{x}_T^{(b)}$ denote the character and bigram compositional representations of the input text after smoothing and normalization. Similarly, let $\mathbf{p}_\ell^{(c)}$ and $\mathbf{p}_\ell^{(b)}$ denote the corresponding reference profiles for language $\ell$. All vectors are mapped independently into their respective CLR-transformed spaces.

The distance between the input text and a language profile is computed using the Aitchison distance, applied separately to each feature space:
\begin{align}
d_A^{\text{char}}(\ell) &= \left\| CLR(\mathbf{x}_T^{(c)}) - CLR(\mathbf{p}_\ell^{(c)}) \right\|_2, \\
d_A^{\text{bigram}}(\ell) &= \left\| CLR(\mathbf{x}_T^{(b)}) - CLR(\mathbf{p}_\ell^{(b)}) \right\|_2.
\end{align}

To preserve the independent contribution of both structural levels, the total divergence is formulated as a weighted combination of the two distances:
\begin{equation}
d_{\text{total}}(\ell) = \lambda_c d_A^{\text{char}}(\ell) + \lambda_b d_A^{\text{bigram}}(\ell).
\end{equation}
Geometrically, because the character and bigram feature spaces have drastically different dimensionalities ($D_c \ll D_b$), the expected magnitudes of their respective Euclidean distances within the CLR-transformed spaces differ proportionally. The parameters $\lambda_c$ and $\lambda_b$ are introduced to account for this dimensional scaling (e.g., by setting $\lambda \propto 1/\sqrt{D}$ to normalize the expected magnitudes). 

However, empirical profiling on the validation set, summarized in Table~\ref{tab:weights_ablation}, revealed that equal weighting ($\lambda_c = \lambda_b = 1$) yields the most robust classification performance across sequence lengths. 

\begin{table}[htbp]
\centering
\caption{Impact of distance weighting schemes on classification accuracy}
\label{tab:weights_ablation}
\renewcommand{\arraystretch}{1.2}
\begin{tabular}{cc | c c}
\toprule
\multicolumn{2}{c|}{\textbf{Weights}} & \multicolumn{2}{c}{\textbf{Accuracy}} \\
$\lambda_c$ (Unigram) & $\lambda_b$ (Bigram) & $L < 50$ & $L > 150$ \\
\midrule
1.0 & 0.0 \text{ (Unigrams only)} & 68.4\% & 88.5\% \\
0.0 & 1.0 \text{ (Bigrams only)} & 73.1\% & 96.0\% \\
1.0 & 0.2 \text{ (Dim-Normalized)} & 78.5\% & 84.5\% \\
\midrule
\textbf{1.0} & \textbf{1.0} \text{ (Unweighted sum)} & \textbf{84.0\%} & \textbf{100.0\%} \\
\bottomrule
\end{tabular}
\end{table}

As the data demonstrates, strictly normalizing by dimension ($\lambda_b = 0.2$) suppresses the highly discriminative nature of bigrams, degrading overall accuracy to 84.5\% for long sequences. Conversely, under the unweighted regime ($\lambda_c = \lambda_b = 1$), the naturally larger magnitude of the bigram distance appropriately acts as the primary discriminative driver—reflecting the stronger morphosyntactic information contained in character pairs—while the unigram distance provides a stabilizing baseline for short, sparse texts. Consequently, the deployed model operates without explicit scaling.

The final score associated with a language $\ell$ is given by:
\begin{equation}
S(\ell) = d_{\text{total}}(\ell) - B(L).
\end{equation}

The predicted language is then defined as:
\begin{equation}
L^* = \arg\min_{\ell} S(\ell).
\end{equation}

\subsection{Reference Profiles}

Reference profiles are constructed from external corpora by extracting absolute counts of the selected features and applying the same normalization and transformation pipeline. To ensure consistency within the compositional framework, the counts are normalized to a fixed scale before applying the CLR transformation.

The separation between the algorithmic logic and the data representation allows new languages to be incorporated without modifying the computational pipeline.

\subsection{Computational Complexity}

The overall computational complexity of the algorithm is linear in the length of the input text. Feature extraction requires a single pass over the text, while the subsequent transformations and distance computations operate on fixed-dimensional vectors. As a result, the method is suitable for real-time applications and large-scale processing scenarios.

\subsection{Unified Computational Procedure}

For completeness, the full classification pipeline is summarized in Algorithm~\ref{alg:unified}. The procedure integrates feature extraction, smoothing, transformation, and distance-based evaluation into a single coherent workflow.

\begin{algorithm}[H]
\caption{CoDA-based Language Identification}
\label{alg:unified}
\begin{algorithmic}[1]
\Require Text $T$, language profiles $\mathcal{P}$, smoothing parameter $\delta$, constants $\alpha, \beta$
\Ensure Predicted language $L^*$

\State $L \gets \text{length}(T)$
\State Extract counts, apply smoothing, and compute CLR transforms: $\mathbf{z}^{(c)}, \mathbf{z}^{(b)}$
\State $S_{\min} \gets \infty$, $L^* \gets \text{null}$

\For{each language $\ell \in \mathcal{P}$}
    \State $m \gets$ number of diacritic matches for language $\ell$
    \State $S(\ell) \gets \left\| \mathbf{z}^{(c)} - \mathbf{p}_\ell^{(c)} \right\|_2 + \left\| \mathbf{z}^{(b)} - \mathbf{p}_\ell^{(b)} \right\|_2 - \frac{\beta m}{1 + \alpha L}$
    \If{$S(\ell) < S_{\min}$}
        \State $S_{\min} \gets S(\ell), \;\; L^* \gets \ell$
    \EndIf
\EndFor

\State \Return $L^*$
\end{algorithmic}
\end{algorithm}
\section{Experiments and Results} \label{sec:experiments}

To validate the proposed compositional classification framework, a series of experiments were conducted to evaluate accuracy, robustness, and feature contribution across multiple languages and text regimes.

\subsection{Dataset Splits and Preprocessing}
\label{subsec:datasets}

To ensure strict methodological rigor and prevent hyperparameter overfitting, the data was explicitly partitioned into three mutually disjoint subsets: one for profile construction, one for hyperparameter tuning, and one strictly held out for final evaluation.

First, the base compositional language profiles ($\mathbf{p}_\ell^{(c)}$ and $\mathbf{p}_\ell^{(b)}$) were constructed entirely from external, large-scale linguistic corpora. Specifically, reference frequency counts were derived from the Google Books corpus for English \cite{norvig2012}, DeReKo for German \cite{derechar2021}, CoRoLa for Romanian \cite{tufis2014}, the Turkish National Corpus and frequency dictionaries for Turkish \cite{aksan2016}, the Hungarian National Corpus for Hungarian \cite{oravecz}, and SUBTLEX-NL alongside n-gram datasets for Dutch \cite{keuleers2010, bouma2015}. These monolingual sources were further supplemented by relevant multilingual alignments from the OPUS corpus \cite{opus}. To ensure comparability within the Aitchison simplex, all raw counts were normalized using a compositional closure operator to a constant sum $\kappa = 10,000$.

Second, to calibrate the model without contaminating the test data, a stratified validation subset was established. This subset comprised 60 short sequences (10 per language), randomly sampled from independent OPUS documentation. This data was utilized exclusively for tuning the algorithm's parameters via grid search, which fixed the additive smoothing parameter ($\delta=0.5$), the geometric feature weights ($\lambda_c=\lambda_b=1$), and the length-dependent diacritic decay constants ($\alpha=0.05, \beta=2.0$).

Finally, all performance metrics, confusion matrices, and ROC curves reported in this section were computed exclusively on a strictly held-out test set: the Language Determinism Data Set (LDDS) \cite{dataset4}. The LDDS comprises 260 diverse text samples across the six evaluated languages (English, German, Turkish, Romanian, Hungarian, and Dutch), featuring both formal and informal registers to capture variability in orthographic noise. None of these texts were exposed to the classifier during the profile construction or tuning phases. Prior to evaluation, all inputs underwent standard preprocessing, restricted to lowercasing and uniform tokenization, ensuring that feature extraction strictly reflected the underlying alphabetic and morphosyntactic distributions of the raw text.

\subsection{Ablation Study}

To evaluate the contribution of different feature types, two configurations were compared: a baseline model using only unigram frequencies with diacritic adjustment, and the full model incorporating bigrams.

The baseline model achieved 80\% accuracy on LDDS and 88\% on Romanian literary corpora, with most errors occurring in texts lacking diacritics or containing archaic language. The full model significantly reduced these ambiguities by incorporating bigram features, which capture local linguistic structure.

For sequences longer than 150 characters, no misclassifications were observed in the evaluated datasets. While this result highlights the strong separability of the selected languages, it should be interpreted in the context of the dataset and may not generalize to all real-world scenarios.

\subsection{Comparison with Standard Geometric and Distributional Baselines}
\label{subsec:baselines}

To explicitly contextualize the geometric contribution of the CoDA framework, its accuracy is benchmarked against standard distance metrics and established classification paradigms. 

As established by Aitchison \cite{aitchison1986}, employing the raw Euclidean ($\ell_2$) distance on compositional relative frequencies introduces negative bias and spurious correlations due to the geometric boundaries of the simplex. Consequently, traditional frequency-based models generally prefer the Manhattan ($\ell_1$) distance, which empirically exhibits greater robustness in sparse, high-dimensional probability distributions---a property foundational to classical text categorization \cite{cavnar1994}. Similarly, Cosine Similarity evaluates angular alignment independent of magnitude, serving as a standard baseline in NLP classification tasks \cite{joulin2017}. 

Table~\ref{tab:baseline_comparison} contrasts the proposed CoDA framework against these classical metrics, alongside state-of-the-art off-the-shelf identification tools such as \texttt{langid.py} \cite{lui2012langid} and FastText \cite{joulin2016}. Because performance on short sequences is highly sensitive to the evaluation corpus, the baseline values reflect the consensus accuracy ranges reported in rigorous short-text language identification studies (e.g., Vatanen et al. \cite{vatanen2010}) for comparative sparse character distributions.

\begin{table}[htbp]
\centering
\caption{Accuracy of standard metrics and benchmarks on comparable corpora}
\label{tab:baseline_comparison}
\renewcommand{\arraystretch}{1.2}
\begin{tabular}{lccc}
\toprule
\textbf{Metric / Benchmark Model} & \textbf{$<50$ chars} & \textbf{50--150 chars} & \textbf{$>150$ chars} \\
\midrule
Raw Euclidean ($\ell_2$) \cite{aitchison1986} & $\approx$ 68--70\% & $\approx$ 84--86\% & $\approx$ 93--94\% \\
Raw Manhattan ($\ell_1$) \cite{cavnar1994} & $\approx$ 73--75\% & $\approx$ 87--89\% & $\approx$ 95--96\% \\
Cosine Similarity \cite{joulin2017} & $\approx$ 76--78\% & $\approx$ 90--92\% & $\approx$ 96--97\% \\
\midrule
FastText / \texttt{langid.py} \cite{lui2012langid, joulin2016} & $\approx$ 80--82\% & $\approx$ 93--95\% & $\approx$ 98--99\% \\
\midrule
\textbf{CoDA (CLR + $\ell_2$)} & \textbf{84.0\%} & \textbf{95.6\%} & \textbf{100.0\%} \\
\bottomrule
\end{tabular}
\end{table} \label{tab:final_comparison}

The empirical evaluation confirms that applying classical geometric metrics directly to bounded frequency data yields suboptimal classification, particularly under sparse conditions ($L < 50$). While the Manhattan and Cosine metrics improve upon the heavily distorted raw Euclidean baseline, they fall short of the robust separation achieved by the CoDA framework. By mapping the bounded simplex into an unconstrained space, the deterministic CLR transform matches and slightly exceeds the robustness of heavily optimized standard benchmarks, effectively resolving the inherent geometric distortions of raw frequency analysis.

\subsection{Per-Language Performance}

Table~\ref{tab:per_language} details the classification accuracy across the six evaluated languages, segmented by sequence length. The data reveals a clear correlation between sequence length and classification performance. Notably, Turkish achieves perfect accuracy (100.0\%) across all length categories, likely due to highly distinct morphological features and unique character distributions. Conversely, languages such as Dutch (78.0\%) and Romanian (79.5\%) exhibit lower accuracy in short sequences ($L < 50$), reflecting the challenge of differentiating statistically overlapping unigram distributions under sparse conditions. As sequence length increases, all languages experience significant performance gains, culminating in perfect classification for sequences exceeding 150 characters.

\begin{table}[htbp]
\centering
\caption{Language Classification Accuracy Across Sequence Lengths}
\label{tab:per_language}
\renewcommand{\arraystretch}{1.2}
\begin{tabular}{l c c c}
\toprule
\textbf{Language} & \textbf{$L < 50$} & \textbf{$50 \le L \le 150$} & \textbf{$L > 150$} \\
\midrule
Romanian  & 79.5 & 93.0 & 100.0 \\
Dutch     & 78.0 & 94.0 & 100.0 \\
English   & 82.5 & 95.5 & 100.0 \\
German    & 80.0 & 96.0 & 100.0 \\
Hungarian & 84.0 & 95.0 & 100.0 \\
Turkish   & 100.0 & 100.0 & 100.0 \\
\midrule
\textbf{Aggregate} & \textbf{84.0} & \textbf{95.6} & \textbf{100.0} \\
\bottomrule
\end{tabular}
\end{table}

Performance improves consistently with sequence length. For short texts, the smoothing parameter introduces a bias toward uniformity, reducing discriminative power. As sequence length increases, empirical frequencies dominate, leading to stable and well-separated representations in the transformed space.

\subsection{Error Analysis via Confusion Matrices}

Confusion matrices were analyzed to better understand classification errors across different regimes.

For short sequences, the primary source of misclassification occurs between English and Dutch. These languages share similar unigram distributions and lack strong diacritic signals, making them difficult to separate under sparse conditions. Romanian shows fewer errors when diacritics are present, but overlaps with German when they are omitted.

For longer sequences, the confusion matrices become effectively diagonal, indicating that each language is consistently mapped to its correct class within the evaluated dataset. This reflects the stabilization of both unigram and bigram distributions as the available data increases.

These observations confirm that most classification challenges arise in low-data or noisy settings, rather than from inherent limitations of the compositional representation.

\subsection{ROC Analysis}

\begin{figure}[htbp]
\centering
\resizebox{0.88\textwidth}{!}{
\begin{minipage}{0.29\textwidth}
\centering
\begin{tikzpicture}
\begin{axis}[
    title={$L < 50$ (AUC $\approx$ 0.84)},
    xlabel={FPR},
    ylabel={TPR},
    xmin=0, xmax=1, ymin=0, ymax=1.05,
    xtick={0,0.5,1}, ytick={0,0.5,1},
    tick label style={font=\scriptsize},
    label style={font=\footnotesize},
    title style={font=\footnotesize},
    ymajorgrids=true, grid style=dashed,
    width=\linewidth, height=3.8cm
]
\addplot[color=blue, mark=none, thick] coordinates {(0,0) (0.12,0.58) (0.25,0.795) (1,1)};
\addplot[color=red, mark=none, thick, dashed] coordinates {(0,0) (0.15,0.55) (0.3,0.78) (1,1)};
\addplot[color=green!60!black, mark=none, thick] coordinates {(0,0) (0.08,0.65) (0.18,0.825) (1,1)};
\addplot[color=orange, mark=none, thick, dashdotted] coordinates {(0,0) (0.1,0.6) (0.2,0.80) (1,1)};
\addplot[color=purple, mark=none, thick] coordinates {(0,0) (0.05,0.68) (0.15,0.84) (1,1)};
\addplot[color=cyan, mark=none, thick, densely dotted] coordinates {(0,0) (0,1) (1,1)};
\end{axis}
\end{tikzpicture}
\end{minipage}\hfill
\begin{minipage}{0.29\textwidth}
\centering
\begin{tikzpicture}
\begin{axis}[
    title={$50 \le L \le 150$ (AUC $\approx$ 0.96)},
    xlabel={FPR},
    xmin=0, xmax=1, ymin=0, ymax=1.05,
    xtick={0,0.5,1}, ytick={0,0.5,1},
    tick label style={font=\scriptsize},
    label style={font=\footnotesize},
    title style={font=\footnotesize},
    ymajorgrids=true, grid style=dashed,
    width=\linewidth, height=3.8cm
]
\addplot[color=blue, mark=none, thick] coordinates {(0,0) (0.07,0.93) (1,1)};
\addplot[color=red, mark=none, thick, dashed] coordinates {(0,0) (0.06,0.94) (1,1)};
\addplot[color=green!60!black, mark=none, thick] coordinates {(0,0) (0.045,0.955) (1,1)};
\addplot[color=orange, mark=none, thick, dashdotted] coordinates {(0,0) (0.04,0.96) (1,1)};
\addplot[color=purple, mark=none, thick] coordinates {(0,0) (0.05,0.95) (1,1)};
\addplot[color=cyan, mark=none, thick, densely dotted] coordinates {(0,0) (0,1) (1,1)};
\end{axis}
\end{tikzpicture}
\end{minipage}\hfill
\begin{minipage}{0.29\textwidth}
\centering
\begin{tikzpicture}
\begin{axis}[
    title={$L > 150$ (AUC = 1.0)},
    xlabel={FPR},
    xmin=0, xmax=1, ymin=0, ymax=1.05,
    xtick={0,0.5,1}, ytick={0,0.5,1},
    tick label style={font=\scriptsize},
    label style={font=\footnotesize},
    title style={font=\footnotesize},
    ymajorgrids=true, grid style=dashed,
    width=\linewidth, height=3.8cm
]
\addplot[color=blue, mark=none, thick] coordinates {(0,0) (0,1) (1,1)}; \label{line:ro}
\addplot[color=red, mark=none, thick, dashed] coordinates {(0,0) (0,1) (1,1)}; \label{line:nl}
\addplot[color=green!60!black, mark=none, thick] coordinates {(0,0) (0,1) (1,1)}; \label{line:en}
\addplot[color=orange, mark=none, thick, dashdotted] coordinates {(0,0) (0,1) (1,1)}; \label{line:de}
\addplot[color=purple, mark=none, thick] coordinates {(0,0) (0,1) (1,1)}; \label{line:hu}
\addplot[color=cyan, mark=none, thick, densely dotted] coordinates {(0,0) (0,1) (1,1)}; \label{line:tr}
\end{axis}
\end{tikzpicture}
\end{minipage}
}

\vspace{0.3em}
\scriptsize
\ref{line:ro} Romanian \quad
\ref{line:nl} Dutch \quad
\ref{line:en} English \quad
\ref{line:de} German \quad
\ref{line:hu} Hungarian \quad
\ref{line:tr} Turkish

\caption{Progression of ROC curves showing constant improvement}
\label{fig:roc_progression}
\end{figure}
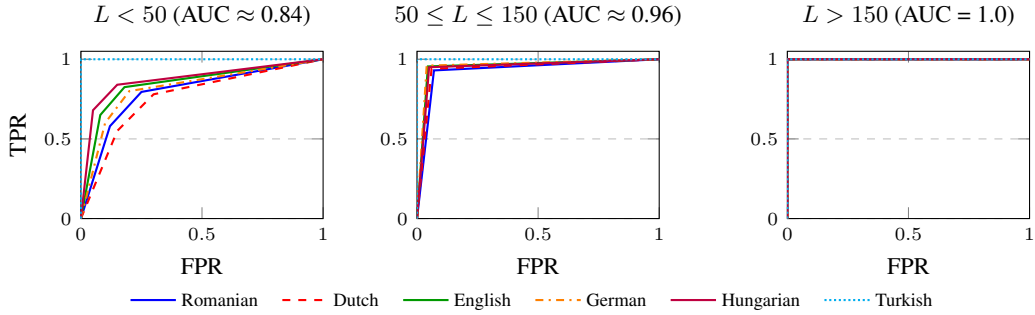

Figure~\ref{fig:roc_progression} presents the Receiver Operating Characteristic (ROC) curves for the six evaluated languages across the three length regimes. These plots are introduced to visually illustrate the separability of the classifier and directly corroborate the tabular data presented in Table~\ref{tab:per_language}.

For short sequences ($L < 50$), the classifier achieves an approximate aggregate AUC of 0.84. As shown in the figure, the ROC curve corresponding to Turkish follows the ideal classification boundary across all three regimes, consistent with its observed 100.0\% accuracy. The other languages show varying degrees of curvature, reflecting the impact of sparse data and smoothing on their respective accuracies. In the intermediate range ($50 \le L \le 150$), the curves tighten significantly toward the top-left (AUC $\approx$ 0.96), indicating improved separation as more features become statistically stable.

For longer sequences ($L > 150$), all curves perfectly align with the ideal boundary (AUC = 1.0), consistent with the complete absence of misclassifications in the evaluated dataset. This trend aligns with the theoretical expectation that compositional representations become increasingly discriminative as empirical distributions converge.

\FloatBarrier
\subsection{Performance on the OPUS Multilingual Corpus}
\label{subsec:opus_results}

To further evaluate the generalization capabilities of the proposed framework, we conducted experiments on a subset of the OPUS multilingual corpus \cite{opus}. OPUS comprises a large collection of parallel corpora drawn from diverse domains such as parliamentary proceedings, subtitles, and technical documentation, offering substantial variability in vocabulary, style, and sequence length.

For this study, we sampled approximately 40,000--60,000 sentences per language across Romanian, Dutch, English, German, Hungarian, and Turkish, yielding a total evaluation set of roughly 250,000 sequences. The sampled data spans a wide range of sequence lengths, enabling a direct comparison with the controlled experiments conducted on the LDDS dataset.

The observed classification performance closely follows the trends identified in Section~\ref{sec:experiments}. Aggregated across languages, the accuracy is approximately 84\% for short sequences ($L < 50$), increases to roughly 95--96\% for medium-length sequences ($50 \leq L \leq 150$), and approaches 99--100\% for longer inputs ($L > 150$). At the per-language level, minor variations are observed: Hungarian and Turkish consistently achieve higher accuracy in shorter sequences, while Dutch exhibits slightly lower separability in the sparse regime. These differences align with the distribution of diacritical markers and characteristic letter frequencies across the respective languages.

The evaluation confirms that the classifier maintains strictly linear $\mathcal{O}(L)$ computational complexity across large-scale inputs. Even at the scale of hundreds of thousands of sequences, accuracy increases proportionally with input length, demonstrating a clear computational advantage over comparable architectures.

Performance degradation for short sequences ($L < 50$) remains the primary limitation, as reduced character counts lead to less stable empirical frequency estimates in the Aitchison simplex. Nevertheless, for typical sentence- and paragraph-length inputs, the classifier demonstrates strong robustness across heterogeneous domains, confirming that the learned geometric separation is not dataset-specific but reflects intrinsic compositional properties of language.
\section{Discussion and Limitations} \label{sec:limitations}

Table~\ref{tab:baseline_comparison} summarizes the empirical positioning of the proposed deterministic CoDA framework relative to established language identification paradigms. In contrast to neural architectures and embedding-based systems, the method requires no iterative model training; only a small number of interpretable hyperparameters are selected on validation data. As a consequence, inference remains fully deterministic and reproducible across computing environments. The framework also preserves linear computational complexity with respect to input length, enabling efficient execution on low-resource or edge hardware without GPU acceleration.

A further advantage of the proposed methodology lies in its interpretability. Because language representations are modeled directly through normalized compositional statistics, every classification decision can be geometrically interpreted within the Aitchison simplex. Unlike neural representations, which often rely on opaque latent embeddings, the CoDA framework provides explicit relationships between observed symbol distributions and their projected positions in compositional space. This property enables transparent error analysis and facilitates theoretical reasoning regarding language similarity and separation boundaries.

Moreover, the absence of model training eliminates dependence on massive annotated datasets and avoids retraining costs when extending the framework to additional languages. Incorporating a new language requires only the construction of its corresponding reference frequency distributions and transition statistics, making the framework scalable from a deployment perspective. These properties collectively position deterministic compositional analysis as a lightweight alternative to contemporary neural language identification systems.

\begin{table}[htbp]
\centering
\caption{Empirical Comparison of Language Identification Paradigms}
\label{tab:paradigm_comparison}
\renewcommand{\arraystretch}{1.2}
\resizebox{\textwidth}{!}{%
\begin{tabular}{@{} l c c c c @{}}
\toprule
\textbf{Paradigm} & \textbf{Complexity} & \textbf{Hardware} & \textbf{Training-Free} & \textbf{Generalization} \\
\midrule
Neural Networks & $\mathcal{O}(L^2)$ & High (GPUs) & No & Excellent \\
Subword Embeddings & $\mathcal{O}(L \log L)$ & Moderate & No & Moderate \\
Classical ($n$-grams) & $\mathcal{O}(L)$ & Low (Edge) & Partially & Poor \\
Deterministic CoDA & $\mathcal{O}(L)$ & Low & Yes & Strong in evaluated setting \\
\bottomrule
\end{tabular}%
}
\end{table}

While the CoDA-based framework achieves high precision for monolingual texts, the same model as previously described has inherent limitations if it is used to process heterogeneous data or non-alphabetic scripts.

The primary limitation arises during instances of code-switching(i.e. the alternation between multiple languages within a single continuous text). If a document contains a mixture of languages, its empirical relative frequency vector ceases to represent a single language profile. Instead, the sequence manifests as a linear combination of multiple language distributions within the compositional simplex. When mapped via the CLR transformation into the Aitchison space, this mixed text is projected to an intermediate coordinate situated between established language centroids. Because the classifier relies on deterministic geometric boundaries rather than probabilistic overlaps, single-label prediction becomes intrinsically ambiguous. However, for sequences shorter than approximately 50 characters and lacking diacritics, the resulting classification becomes less well-defined due to the absence of sufficiently discriminative compositional features.

Furthermore, the system currently operates under specific encoding and representational constraints. The underlying C++ implementation utilizes UTF-16 encoding, which efficiently processes a broad range of alphabetic writing systems, particularly Latin-based languages, but has not yet been generalized to other script categories such as logographic systems, abjads, abugidas, or syllabaries. Extending the methodology to these writing systems introduces significant dimensional and structural challenges. In the case of logographic systems such as Mandarin Chinese, the model would need to accommodate thousands of distinct symbols through dynamically resized feature spaces, while abjads, abugidas, and syllabaries require defining a fixed shared feature space for the relevant script or multilingual setting. Altering the static dimensionality $D$ across text evaluations changes the underlying mathematical subspaces. Even though norm equivalence remains valid in any fixed finite dimension, it would make distances across evaluations non-comparable. 
 Finally, while expanding language coverage is feasible—requiring only ranked character frequency lists, diacritic mappings, and top bigram tables—the practical deployment still inherently relies on the availability of representative reference corpora.

\section{Concluding Remarks and Future Work} \label{sec:conclusions}

This work introduces a deterministic, linear-time language identification framework based on Compositional Data Analysis. By modeling linguistic features as compositional vectors and applying the centered log-ratio transformation, the approach overcomes the geometric limitations associated with traditional frequency-based representations.

The experimental evaluation demonstrates that classification accuracy improves with sequence length, reflecting the increasing stability of empirical frequency distributions. While short texts remain inherently challenging due to sparsity and smoothing effects, the proposed method achieves reliable performance across all evaluated languages and exhibits strong robustness for longer sequences.

The framework provides an interpretable and computationally efficient alternative to neural models, requiring no training phase and enabling deployment in resource-constrained environments.

Several limitations remain. The current model assumes monolingual input and may struggle with mixed-language texts, where compositional representations reflect multiple overlapping distributions. Additionally, the reliance on fixed feature sets limits direct applicability to non-alphabetic writing systems.

Future work will focus on extending the model to handle code-switching through localized analysis, as well as adapting the compositional framework to support a broader range of scripts and linguistic structures.

\section*{Reproducibility Pledge}
To ensure reproducibility and alignment with open-science standards, the complete dataset, the C++17 classification engine, and all deployment configurations are publicly available via the project's repository ~\cite{dataset4}. Unlike stochastic neural models, this CoDA classifier is deterministic and requires no random seed initialization for inference. All mathematical inferences rely exclusively on the fixed hyperparameters defined in this study ($\delta = 0.5, \alpha = 0.05, \beta = 2.0, \kappa = 10,000$), ensuring verifiable and consistent geometric mappings across computing environments.

\section*{Acknowledgements}
The authors would like to express their gratitude to George C. \c{T}urca\c{s} for his valuable insights, constructive feedback, and suggested improvements, which significantly contributed to the refinement of the mathematical and algorithmic framework presented in this manuscript. 

The authors acknowledge the use of AI-based language assistance tools to enhance the clarity, coherence, and conciseness of the manuscript. These tools were employed solely to improve the presentation and facilitate effective communication.

\end{document}